\documentclass[11pt,a4paper]{article}

\usepackage[hyperref]{emnlp2021}
\usepackage{times}
\usepackage{latexsym}
\usepackage[T1]{fontenc}
\usepackage[utf8]{inputenc}
\usepackage{microtype}
\usepackage{tabu}
\usepackage{amsmath}
\usepackage{booktabs}
\usepackage{makecell}
\usepackage{tikz}
\usepackage{xspace}
\usepackage{enumitem}
\usepackage{tikz}

\setlist{nosep}

\title{Are Transformers a Modern Version of ELIZA? Observations on French Object Verb Agreement}

 \author{Bingzhi Li \and Guillaume Wisniewski \and Beno\^it Crabb\'e\\
  Université de Paris, LLF, CNRS \\
  75\,013 Paris, France \\
  {\tt bingzhi.li@etu.u-paris.fr} \\
  {\tt \{guillaume.wisniewski,benoit.crabbe\}@u-paris.fr} \\}

\newcommand{\alert}[1]{\textcolor{blue}{\textbf{#1}}}

\begin{document}
\maketitle

\begin{abstract}
  Many recent works have demonstrated that unsupervised sentence
  representations of neural networks encode syntactic information by
  observing that neural language models are able to predict the
  agreement between a verb and its subject.  We take a critical look
  at this line of research by showing that it is possible to achieve
  high accuracy on this agreement task with simple surface heuristics,
  indicating a possible flaw in our assessment of neural networks'
  syntactic ability. Our fine-grained analyses of results on the
  long-range French object-verb agreement show that contrary to LSTMs,
  Transformers are able to capture a non-trivial amount of grammatical
  structure.
\end{abstract}

\section{Introduction}

The long distance agreement task is one of the most popular method to
assess neural networks (NN) ability to encode syntactic information:
\newcite{linzen-etal-2016-assessing} showed that LSTMs are able to
predict the subject-verb agreement in English and has initiated a very
active line of research. Since then, many studies have generalized
this observation to other languages
\cite{gulordava-etal-2018-colorless}, other models such as
Transformers~\cite{goldberg19,jawahar-etal-2019-bert} or have
identified possible confounding factors that could distort the stated
conclusions
\cite{gulordava-etal-2018-colorless,marvin-linzen-2018-targeted}. All
of these studies show that NN are able to learn a `substantial amount'
of syntactic information~\cite{belinkov-glass-2019-analysis}.

In this work, we propose to take an alternative look at these results
by studying whether neural networks are able to predict the correct
form of a verb because they are able to build an abstract, high-level
(maybe hierarchical) sentence
representation~\cite{giulianelli2018under,lakretz2019emergence} or
solely because they capture surface statistical regularities, as
suggested by several recent
work~\cite{sennhauser-berwick-2018-evaluating,chaves-2020-dont,li-wisniewski-2021-neural
}. Overall, this set of results questions one of the most fundamental
assumption in linguistics~\cite{lakretz21}, namely that a sentence has
a recursive structure~\cite{everaert2015}: while LSTMs with proper
parametrization can model context-free patterns 
\cite{
  suzgun:2019}, Transformers
are essentially feed forward models relying on a large number of
attention heads. Consequently, they are, in theory, not adapted to model
hierarchical syntactic patterns \cite{hahn:2020} and explaining their
capacity to predict accurately syntactic agreement patterns remains an
open issue.

We bring new light on this problematic by identifying simple
heuristics (\textsection\ref{sec:expe}) that can be used to correctly predict
verbal agreement, pushing further the observation of
\newcite{kuncoro2018perils} that a simple rule can provide highly
accurate results on the task. Using our extended set of heuristics, we
identify sentences for which predicting the correct verb form requires
a more abstract representation of the sentence. By comparing models'
performance on these examples, we show that contrary to LSTMs,
Transformers perform consistently well in these critical
cases. 

\section{Test Set for French Object Past-Participle Agreement\footnote{The   code   of   all   our   experiments  as well  as  the  corpora  we  used  in  this  work can  be  downloaded  from \url{https://gitlab.huma-num.fr/bli/syntactic-ability-nlm}.}\label{sec:background}}

\begin{figure*}
\tikzstyle{every picture}+=[remember picture,inner xsep=0,inner ysep=0.25ex]

\scalebox{0.97}{
  \begin{tabu}{cllllllllll}
  (1) &Le & nombre & d' &  \tikz[baseline=(node1.base)]\node (node1) {offres}; & \tikz[baseline=(node2.base)]\node (node2){\alert{ que} }; & le & directeur & a& \alert{acceptées}& \\
  \rowfont{\tiny}    &The-DET-Sg & number-N-M-Sg & of-ADP &  offers-N-\alert{F-Pl} & that-PRON & the-DET-3Sg & director-N-M-Sg &has-AUX-3Sg & accepted-PP-\alert{F-Pl} &\\
  & \multicolumn{9}{l}{The number of offers that the director has accepted...} \\\\
  
  (2) &Les &  \tikz[baseline=(node3.base)]\node (node3){offres$_{\textrm{\textcolor{orange}{h1}}}$}; & \tikz[baseline=(node4.base)]\node (node4){\alert{que}}; & les& directeurs$_{\textrm{\textcolor{orange}{h2}}}$ &  ont$_{\textrm{\textcolor{orange}{h3}}}$ & \alert{acceptées} &...&  \\
  \rowfont{\tiny}&The-DET-Pl & offers-N-\alert{F-Pl} & that-PRON & the-DET-Pl &directors-N-M-{\bf Pl}& have-AUX-3{\bf Pl} & accepted-PP-\alert{F-Pl} &...&   \\
  & \multicolumn{9}{l}{The offers that the directors have accepted... } \\  
  \end{tabu}}
  \caption{Examples of object-verb agreement in French. The past
    participle in the relative clause (in blue) has to agree in gender
    and in number with its object (also in blue) when the latter is
    placed before the verb. To predict the agreement the model has to identify the antecedent of the relative pronoun (dashed arrow)\label{fig:ex_agreement}}
    
    \begin{tikzpicture}[overlay]
    \draw[-latex,dashed] (node2.north) to[bend right] (node1.north);
    \draw[-latex,dashed] (node4.north) to[bend right] (node3.north);

\end{tikzpicture}
\end{figure*}
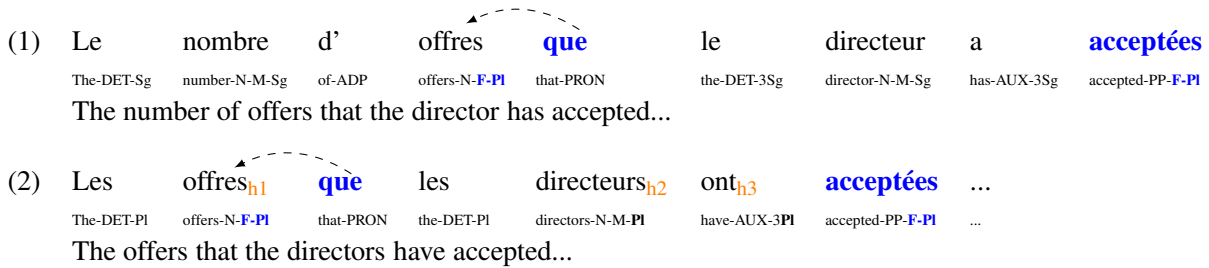

We focus on the object-verb agreement (i.e.\ object past-participle agreement) in French: agreement in number
and gender occurs between the object and the past participle when the
latter is used with the auxiliary \textit{avoir} (to have) and the object is
located before the verb. As shown in Figure~\ref{fig:ex_agreement},
this is, for instance, the case for past participles in object
relatives.  When agreement is required, a \texttt{-s} suffix (resp.\
\texttt{-e}) has to be added to past participles for plural object
(resp.\ feminine).

To predict the past participle agreement in object relatives, a model
has to identify the object relative pronoun, its antecedent and the
auxiliary. It has also to ignore the effect of attractors (nouns with
misleading agreement features) occurring between the object and the
past participle.  Compared to the subject-verb agreement, the French
object past participle agreement is more difficult as the target verb
form depends on a noun that is never adjacent to the verb. The
auxiliary \textit{avoir} before the target verb could also be an
attractor. \footnote{See example(1) in Figure~\ref{fig:ex_agreement},
  \textit{a} (has\_3Sg) could be a number attractor for target verb
  \textit{acceptées}(accepted\_Pl)} 
%

We restrict ourselves to the number agreement between object and past
participle in the case of object relatives to (1) design reasonably
simple patterns that can be easily extracted automatically from raw
texts, (2) extract a sufficiently large number of representative
examples and (3) reduce the importance of the anaphoric resolution
problem.  These restrictions allow us to carry out a fine-grained
analysis of NN  ability to extract syntactic generalizations
from non-annotated corpora (\textsection\ref{sec:expe}).

\paragraph{Building a Test Set} Sentences used in the number agreement
task are extracted automatically from the 8,638,145 sentences of the
French Gutenberg corpus.\footnote{\url{https://www.gutenberg.org/}} We
use the \textsc{FlauBERT} parser~\cite{le2019flaubert} and the
pretrained French model of \texttt{spaCy}~\cite{spacy} to
automatically parse the sentences of the Gutenberg project. We also
consider the gold annotations of the 45,088 sentences of the French
Universal Dependency treebanks \cite{ud27} to evaluate the impact of
parsing errors on the corpus quality.

We extract examples of object-verb agreement from sentences' syntactic
and morphological annotations using simple rules,\footnote{See
  appendix~\ref{sec:construction} for a full description} resulting in
a corpus of 104 sentences (68\% singular and 32\% plural) extracted
from the UD treebank and of 68,794 sentences (65\% singular and 35\%
plural) extracted from the Gutenberg project. In French, the singular
is identical to the unmarked form of the past participle verbs, making
the frequency statistics unbalanced in favor of singular.

We evaluate the quality of our automatic extraction procedure by
comparing the examples extracted using the gold annotations of UD treebank to those extracted from predicted annotations of UD treebank sentences (generated by \textsc{FlauBERT} and \texttt{spaCy}). 
Our automatic procedure correctly picked up 98\%\footnote{Qualitative
  analysis is in Section~\ref{sec:construction} of the appendix.} of
the object past participle agreement examples.

\section{Language Models \label{sec:models}}

We contrast two types of incremental language models in our
experiments: LSTM models and incremental Transformer models. Both
models are modeling the probability of a sentence $\mathbf{x}$ as:
\begin{equation}
P(\mathbf{x}) = \prod_{i=1}^n P(x_i|x_1\ldots x_{i-1})
\end{equation}
All neural models are trained to compute 
$P(x_i|x_1\ldots x_{i-1})$ and they all use the same generic template:
\begin{align}
    P(x_i|x_1\ldots x_{i-1}) &= \textsc{softmax}(\mathbf{W}_{dec} \mathbf{c}_{i-1}+\mathbf{b})\\
    \mathbf{c}_{i-1} &= \textsc{context}(\mathbf{e}_1\ldots \mathbf{e}_{i-1}) \\
    \mathbf{e}_i&= \mathbf{W}_{enc}\mathbf{x}_i
\end{align}
where $\mathbf{x}_i$ are one-hot word vectors; $\mathbf{W}_{enc}$ and
$\mathbf{W}_{dec}$ are tied parameter matrices, the latter being the
transpose of the former, encoding respectively the word embeddings and
the output layer of the language model.

A context model (\textsc{context}) is either an incremental LSTM or a
Transformer decoder where the sequence of embeddings
$\mathbf{e}_i\ldots \mathbf{e}_n$ is masked (i.e.\ the probability of
the $i$-th word is estimated knowing only the first (i-1) words of
the sentences, contrary to the `standard' Transformer models which
assume that the whole sentence is known). The context vector
$\mathbf{c}$ returned by the context model is either the hidden vector
of the LSTM at step $i-1$ or the vector returned by the upper layer of
the Transformer at step $i-1$.

Our LSTM models use 2 layers while our Transformer language model use
16 layers and 16 heads. Both models are using embeddings of size 768
and are trained on the same data. For Transformers we add positional
embeddings to the word embeddings $\mathbf{e}_i$ using the cosine
scheme and weighting described by \newcite{vaswani2017attention}.
Since all the models use a word-based tokenization and not a subword
tokenizer, we bound the vocabulary to the 50,000 most frequent tokens
found in the training data and use an {\verb+<unk>+} token to encode
the least frequent tokens.

This setting aims to get reasonably fair comparisons between LSTM and
Transformers. To train the models, we extracted raw text from a recent
French Wikipedia dump using
\texttt{WikiExtractor}~\cite{Wikiextractor2015} and then segmented and
tokenized it with the \texttt{Moses}
tokenizer~\cite{koehn2007moses}. We filtered out sentences with more
than 5\% unknown words based on the lemma annotations generated by
\texttt{TreeTagger}~\cite{schmid1999improvements}. Finally, we sampled
a subset containing 100M tokens and split it into training, validation
and test sets with a standard 8:1:1 proportion.

\section{Experimental Results \label{sec:expe}}

\begin{table}
  \centering
  \scalebox{.8}{
  \begin{tabular}{lllcc}
    \toprule
    \multicolumn{2}{l}{corpus} & \makecell{size \\ {\scriptsize in sentences}} & LSTMs & Transformers \\
    \midrule
    \multicolumn{5}{l}{\textit{Original Test Set}} \\
    \phantom{ab} & overall & 68,497 & 80,8$_{\pm 1.2}$   & 93.5 $_{\pm 1.4}$ \\\cline{2-5}
    \phantom{ab} & 4 heuristics &32,311 &96.4 $_{\pm 0.6}$ & 99.0 $_{\pm 0.4}$ \\
    \phantom{ab} & 3 heuristics & 13,222 & 84.0 $_{\pm 1.7}$ & 95.1 $_{\pm 1.5}$\\
    \phantom{ab} & 2 heuristics & 8,869 & 66.5 $_{\pm 2.7}$& 89.5 $_{\pm 2.3}$\\    
    \phantom{ab} & 1 heuristic & 10,946&55.7 $_{\pm 3.5}$& 84.2 $_{\pm 3.0}$\\
    \phantom{ab} & 0 heuristic & 3,149&34.9 $_{\pm 6.8}$ & 74.1 $_{\pm 4.1}$\\
    \midrule
    \multicolumn{4}{l}{\textit{Permuted Test Set}} \\
     & overall & 68,497 & 69.0 $_{\pm 0.6}$  & 70.4 $_{\pm 1.0}$     \\\cline{2-5}
     & 4 heuristics &32,311 &87.0 $_{\pm 1.2}$ & 88.0 $_{\pm 0.8}$\\
     & 3 heuristics & 13,222 & 73.6 $_{\pm 0.6}$& 73.9 $_{\pm 0.9}$\\
     & 2 heuristics & 8,869 & 52.0 $_{\pm 0.3}$& 54.8 $_{\pm 1.8}$\\    
     & 1 heuristic & 10,946&35.3 $_{\pm 0.3}$& 37.4  $_{\pm 1.5}$ \\
     & 0 heuristic  & 3,149&30.2 $_{\pm 0.4}$& 32.6 $_{\pm 1.2}$\\
    \bottomrule
  \end{tabular}}
\caption{Accuracy achieved by LSTMs and Transformers on the
  object-verb agreement task for the \textit{Original} and
  \textit{Permuted} test sets. Results are averaged over the three
  best models in terms of the validation perplexity for each
  architecture\label{tab:results}}
\end{table}

In our experiments, following \newcite{linzen-etal-2016-assessing} and
\newcite{gulordava-etal-2018-colorless}, we compare the probabilities
a language model assigns to the singular form of the target participle
and its plural form given a \emph{prefix}.\footnote{The prefix is made
  of words from the beginning of a sentence up to and excluding the
  target past participle} We consider the model has predicted the
agreement correctly if the form with the correct number has a higher
probability than the form with the incorrect number.

Table~\ref{tab:results} reports the accuracy of two types of models
evaluated in this framework.
Even for this difficult task, the models perform,
overall, very well: LSTMs achieve an accuracy of 80.8\%, a
performance similar to the one reported in the
literature.\footnote{For instance, for the subject-verb agreement
  task, \newcite{gulordava-etal-2018-colorless} reported an overall
  accuracy of 81\% for English and \newcite{mueller-etal-2020-cross} of
  83\% for a wide array of constructions in French.}  With an accuracy
of 93.5\%, Transformers perform even better. These preliminary results support
the conclusion, drawn by many works, that neural networks encode
syntactic information.

However, we believe that this conclusion must be taken with great
care: because of confounding factors, a language model could predict
the correct form without actually capturing syntactic information. For
instance, as our test set is unbalanced~(section \ref{sec:background})
a naive model always choosing the singular form of the participle
achieves an accuracy of 65\%, a score that puts into perspective the
performance of LSTMs. More importantly, \newcite{gulordava-etal-2018-colorless} and \newcite{kuncoro2018perils} observed that the
agreement task can be partially solved by collocational information or
a simple heuristic, namely the number of the first noun of the
prefix. In the following, we propose several experiments to strengthen
these first results.

\subsection{Agreement with Surface Heuristics}

Extending the observations of \newcite{kuncoro2018perils}, we
identify four heuristics that a model could adopt to predict the verb's number only from surface
information. Each of these heuristics assumes that the target past
participle agrees systematically in number with:
\begin{enumerate}[label=h\arabic*.]
\item the \emph{first noun} in the prefix;
\item the \emph{last noun} in the prefix;
\item the \emph{last token} in the prefix with a mark of number;
\item the majority number expressed in the prefix.
\end{enumerate} 

The example (2) in Figure~\ref{fig:ex_agreement} illustrates the
tokens that each heuristic relies on to make its decision. These
heuristics are not tailored to the prediction of the object-past
participle agreement in French: they could easily be used to other
agreement tasks in other language. More complicated, task-specific heuristics
could have been designed. We could, for instance, consider the
first noun on the left of the relative pronoun.

Surprisingly enough, as reported in Table~\ref{h_accu}, on our test
set, these heuristics achieve an accuracy between 60.3\% (for h3) and
88.6\% (for h2). These results challenge our previous conclusion: they
show that the ability to predict the correct number of the verb cannot
be used to prove that a model captures abstract syntactic relations,
since a simple surface heuristic outperforms LSTMs and achieves an
accuracy only slightly worse than that of Transformers. On the
contrary, it suggests that NN, like Eliza, only extract and combine
surface patterns to make their
decisions.

\begin{table}
  \begin{tabular}{ll}
    \toprule
    Heuristics & Accuracy \\
    \midrule
    h1: First noun     & 69.5\% \\
    h2: Last noun     & 88.6\%\\
    h3: Last token     & 60.3\%\\
    h4: Majority number     & 70.0\%    \\
    \bottomrule
  \end{tabular}
  \caption{Heuristics' accuracy on French object past participle agreement task \label{h_accu}}
\end{table}

To shed further light on this new perspective, we use these heuristics
to quantify the `difficulty' of the task: for each example of our test
set, we count the number of heuristics that predict the correct form
and consider that the higher this number, the easier the
prediction. We then divide our test set into five different subsets
according to the number of heuristics that a model could rely on to
predict the verb form: the \textit{4 heuristics} group gathers the
`easiest' examples, while examples in the \textit{0 heuristic} group
are the most difficult, for which the choice of the verb number cannot
rely on simple surface heuristics and requires building a more
abstract representation of the sentence. \footnote{Table
  \ref{examples_h} in the appendix describes examples of sentences in
  each of these groups.} 

Table~\ref{tab:results} reports the results achieved by our models
according to the prediction difficulty. The two architectures have a very
different behavior: while they both show high agreement prediction accuracy 
in the simplest case (the \textit{4 heuristics} group),
LSTMs' performance drops sharply with increasing task difficulty: with an
accuracy of only 34.9\% on the most difficult examples (the \textit{0
  heuristic} group), they perform worse than random. On the contrary,
even if Transformers' performance also degrades with increasing task difficulty, they
perform consistently much better on all groups: they are still 
predicting the correct verb number for 74.1\% of the most difficult
examples, suggesting that Transformers are able to extract certain abstract generalizations. 
\begin{table}
  \centering
  \scalebox{.8}{
  \begin{tabular}{lllcc}
    \toprule
    \multicolumn{2}{l}{corpus} & \makecell{size \\ {\scriptsize in sentences}} & LSTMs & Transformers \\
    \midrule
    \multicolumn{5}{l}{\textit{Original Test Set}} \\
    \phantom{ab} & overall & 68,497 & 80,8$_{\pm 1.2}$   & 93.5 $_{\pm 1.4}$ \\\cline{2-5}
    \phantom{ab} & singular &44,599 &96.4 $_{\pm 1.1}$ & 98.9 $_{\pm 0.4}$ \\
    \phantom{ab} & plural & 23,898 & 51.6 $_{\pm 4.7}$ & 83.5 $_{\pm 3.3}$\\

    \midrule
    \multicolumn{5}{l}{\textit{Nonce Test Set}} \\
    \phantom{ab} & overall & 68,497*3 & 78,1$_{\pm 1.2}$   & 92.6 $_{\pm 1.9}$ \\\cline{2-5}
    \phantom{ab} & singular &44,599*3 &93 $_{\pm 2.3}$ & 96.8 $_{\pm 0.9}$ \\
    \phantom{ab} & plural & 23,898*3 & 50.3 $_{\pm 6.8}$ & 84.7 $_{\pm 3.6}$\\
   
    \midrule
    \multicolumn{4}{l}{\textit{Mirror Test Set}} \\
     & overall & 68,497 & 59,8 $_{\pm 2.5}$  & 81.3 $_{\pm 2.7}$     \\\cline{2-5}
     & singular &23,898 &90.6 $_{\pm 1.8}$ & 91.8 $_{\pm 0.7}$\\
     & plural &44,599  & 43.5 $_{\pm 4.5}$& 75.8 $_{\pm 3.8}$\\
    
    \bottomrule
  \end{tabular}}
  \caption{Accuracy achieved by LSTMs and Transformers on different experimental settings, by target verb number and averaged\label{results_by_number}}
\end{table} 
\subsection{Control Experiments}

To corroborate these results and avoid some known pitfalls of the
agreement task, we have performed four control experiments.

\paragraph{Lexical Cues} Following
\newcite{gulordava-etal-2018-colorless}, we convert the original test
set into a nonsensical but grammatically correct test set to ensure
that the model is not using collocational information to choose the
correct form of the verb. \footnote{Generation procedure is detailed
  in Section~\ref{sec:construction} of the appendix.} Results in
Table~\ref{results_by_number} show that for LSTMs
(resp. Transformers), the global accuracy drops from 80.8\%
(resp. 93.5\%) for the original set to 78.1\% (resp. 92.6\%) for the
so-called \textit{nonce} test set. This drop is of the same order of
magnitude as that reported by \newcite{gulordava-etal-2018-colorless},
showing that the lexical or collocational confounds have only a
moderate impact on models' performance in our agreement prediction
task.

\paragraph{Frequency Bias and Imbalanced Data}
Another possible confound identified in this work
(\textsection\ref{sec:background}) results from the imbalance between
classes: most of the past participles in French are singular and 65\% of the target past participle in our test set are singular. 
That is why, as expected, models perform better in predicting singular
form than plural form(\textit{Original Test Set} of Table \ref{results_by_number}): both LSTMs and Transformers predict almost
perfectly singular forms (accuracy: 96.4\% and 98.9\%), but accuracy
on plural verbs drops sharply: LSTMs correctly predict 51.6\% of the
plural forms and Transformers appear to be more robust with an
accuracy of 83.5\%. 

To ensure, a model is not simply memorizing the most frequent form of
a verb, we have generated a \textit{mirror test set} in which each
plural verb is automatically transformed into singular (and
vice-versa) as well as the corresponding object and all its adjective
and pronoun modifiers to make sure that the modified sentence is grammatically correct.

The accuracy of LSTMs and Transformers on the \textit{mirror set} is of 59.8\%
and 81.3\% (Table \ref{results_by_number}). This drop 
suggests that more frequent forms are more likely to be better
predicted, even though Transformers are more robust to the low
frequency bias. Compared to the \textit{nonce} setting, models'
performance is impacted to a much larger degree in \textit{mirror}
setting. We don't have a clear explanation to this surprising
observation, which need to be explored through new experiments.

\paragraph{Distance} Following \newcite{linzen-etal-2016-assessing} we
have examined how models' performance on this agreement task is
affected by distance between the object and the target verb. Results,
reported in Table \ref{distance} in the appendix show that models'
performance decreases slightly as the distance increases, except for
the shortest distance, thus replicating the results of
\newcite{linzen-etal-2016-assessing}.

\paragraph{Word Order} We now test to which extent a model relies on
word order to predict the verb number. We convert each original
example into a scrambled example by randomly permuting its prefix.  As
reported in Table~\ref{tab:results}, despite the fact that the syntax
has been destroyed in shuffled prefixes setting, both models still
achieve high accuracy for the easy examples but achieve worse than
chance accuracy for the \textit{0} and \textit{1 heuristic} groups,
confirming that syntactic information is critical for models to solve
the most difficult cases. For Transformers, the difference in accuracy
between the original and permuted setting on the \textit{0 heuristic}
group extends up to 41.5 percentage points!
These results suggest that Transformers perform significantly better than surface heuristics and
capture a non trivial amount of word order information.

\section{Conclusions}
We ran a fine-grained analysis of NN's syntactic generalization
capabilities in processing French object-verb agreement for
grammatical number, a phenomenon crucially depending on hierarchical
syntactic structure. We designed a new evaluation protocol based on four
shallow heuristics that the models could adopt to perform number
agreement task. Our experiments show that, contrary to LSTMs,
Transformers extract a non trivial amount of syntactic information.

In future work, we will investigate the kind of syntactic information
Transformers are encoding and the relationship between the superficial
heuristics and hierarchical syntactic structure processing in Transformer models. In particular, our results intriguingly
suggest that Transformers rely on word order information to predict
verb agreement, despite the fact that they don't model word order
explicitly beyond marking each word with its absolute-position
embedding. We plan to study this question in future work.

\section*{Acknowledgments}
We sincerely thank the reviewers and Program Chairs for their careful
reviews and insightful comments, which are of great help in improving
the manuscript. This work was granted access to the HPC resources of French Institute for
Development and Resources in Intensive Scientific Computing (IDRIS) under the allocation 2020-AD011012282 and 2021-AD011012408 made by GENCI.

\bibliography{custom,new_anthology}
\bibliographystyle{acl_natbib}

\clearpage
\appendix

\section{Language Models} 
\label{sec:LM} 

\paragraph{Hyperparameters and perplexities} The results reported in the paper are averaged over three best models in terms of the validation perplexity after 40 training epochs for LSTMs and 50 training epochs for Transformer. The detailed information of the top 3  LSTM and Transformer models is described in table~\ref{hyperparameters}. 

For the LSTM models, we  used embeddings of size 768, with 2 layers. The total parameters are 47,900,241 and we explored the following hyperparameters, for a total of 12 combinations:
\begin{enumerate}
\item batch size: 32, 64(only for learning rate 0.0001)
\item dropout rate: 0.0, 0.1, 0.2, 0.3
\item learning rate: 0.001, 0.0001
\end{enumerate} 

For the Transformer models we used embeddings of size 768, with 16 layers, each with 16 heads. The total parameters are 126,674,513. Training was performed with stochastic gradient descent. The initial learning rate was fixed to 0.02 and we used a cosine scheduling on 50 epochs without annealing. The first epoch was dedicated to warmup with a linear incremental schedule for the learning rate. Batches are of size 64 run in parallel on 8 GPUs except for warmup where the size was fixed to 8.  We explored the initial learning rate of 0.01 and 0.02, the dropout rate of 0.0, 0.1, and 0.2, resulting in a total of 6 combinations.

\section{Surface heuristics}
\label{sec:h} 
We defined four heuristics that a model could adopt to predict the verb's number only from surface information. And then we divided the test set into five subsets based on the number of heuristics, Table~\ref{examples_h} describes the examples for each subgroup.

\begin{table*}
  \begin{tabular}{lccccccc}
    \toprule
     & \shortstack{ hidden/ \\ embedding size} & layers & batch size & dropout rate & learning rate& best epoch & ppl \\
    \midrule
    LSTM     & 768 & 2 & 32 & 0.1 & 0.001 & 21 & 40.5\\
         & 768 & 2 & 32 & 0.2 & 0.0001 & 38 & 39.3\\
        & 768 & 2 & 64 & 0.2 & 0.0001 & 36 & 37.9\\
    Transformer     & 768 & 16 & 64 & 0 & 0.02 & 41 & 31.4 \\
         & 768 & 16 & 64 & 0.2 & 0.01 & 50 & 28.5 \\
        & 768 & 16 & 64 & 0.1 & 0.01 & 49 & 28.2 \\
    \bottomrule
  \end{tabular}
  \caption{Hyperparameters and perplexities of top 3 LSTMs and Transformers used in this work \label{hyperparameters}}
\end{table*}

\begin{table*}
\scalebox{.9}{
  \begin{tabular}{llll}
    \toprule
    Subsets & Examples & Heuristics & class\\
    \midrule
    
    4& $_{\textrm{\textcolor{orange}{h4}}}$Les offres$_{\textrm{\textcolor{orange}{h1}}}$ que les directeurs$_{\textrm{\textcolor{orange}{h2}}}$ ont$_{\textrm{\textcolor{orange}{h3}}}$ acceptées...& h1,h2,h3,h4 & Plural  \\
    & \textit {The offers\_\textbf{Pl} that the\_Pl directors\_Pl have\_Pl accepted\_\textbf{Pl} ...} & \\

    3& $_{\textrm{\textcolor{orange}{h4}}}$Le nombre d'offres$_{\textrm{\textcolor{orange}{h2}}}$ qu'ils ont$_{\textrm{\textcolor{orange}{h3}}}$ acceptées...& h2,h3,h4 &Plural    \\
    & \textit {The number\_Sg of offers\_\textbf{Pl} that they\_Pl  have\_Pl accepted\_\textbf{Pl} ...} &    \\

     2& Les offres$_{\textrm{\textcolor{orange}{h1}}}$$_{\textrm{\textcolor{orange}{h2}}}$ qu'il a acceptées... &h1,h2 & Plural    \\
     & \textit {The offers\_\textbf{Pl} that he\_Sg has\_Sg accepted\_\textbf{Pl} ...} &    \\
     
    1& Les offres$_{\textrm{\textcolor{orange}{h1}}}$ que le directeur a acceptées... &h1  & Plural   \\ 
    & \textit {The offers\_\textbf{Pl} that the\_Sg director\_Sg has\_Sg accepted\_\textbf{Pl} ...} &    \\

    0& Le nombre d'offres que le directeur a acceptées... & none &Plural     \\
    & \textit {The number\_Sg of offers\_\textbf{Pl} that the\_Sg director\_Sg has\_Sg accepted\_\textbf{Pl} ...} &    \\

    \bottomrule
  \end{tabular}}
  \caption{Examples of five subsets according to the number of heuristics that a model could rely on to predict the verb form \label{examples_h}}
\end{table*}

\section{Construction of test sets} 
\label{sec:construction}
\paragraph{Extraction procedure} Extraction of the object-verb agreement examples is based on the dependency structure and morphological information of sentences. Concretely, a valid example has to include a
\textsc{Noun} and \textsc{Verb} connected by an \texttt{acl:relcl}
dependency arc as well as a direct object \textit{que} (that); the
auxiliary connected to the target verb has to be \textit{avoir}
(to have). Using the morphological information, we filtered out sentences
in which the noun and the verb do not agree in number and gender as
well as sentences in which not all words from the antecedent to the target(enclosed) 
occur in the language model's
vocabulary. To reduce the importance of anaphoric resolution problems, we have ruled out the complex and ambiguous cases: long distance dependencies (First example in Figure~\ref{fig:ex_test_set}) and coordinated object noun phrase as antecedent case (Second example in Figure~\ref{fig:ex_test_set}). But we didn't exclude the propositional phrase as antecedent case, because there is no ambiguity in determining the antecedent of the relative pronoun, illustrated by the third example in Figure~\ref{fig:ex_test_set}. 
\paragraph{Qualitative evaluation of extraction procedure} Our automatic extraction procedure correctly identified 102 examples from automatically parsed UD treebanks sentences among 104 examples using the gold annotation of French UD treebanks. Our procedure excluded the first missed example by annotating the intervening relative pronoun \textit{que} (that) as conjunction: \textbf {formule} \textbf{qu}’avec un sens de la nuance plus marseillais que britannique, le président de l’académie a \textbf {appliquée} (\textit{\textbf{formula}$_{\textrm{\textcolor{blue}{Fem-Sg}}}$ with a sense of nuance that was more Marseillais than British, \textbf{that} the president of the academy \textbf{applied}$_{\textrm{\textcolor{blue}{Fem-Sg}}}$}). And for the second one: une manière de \textbf{révolution} sur lui-même, qu’il a \textbf{opérée}... (A way of \textbf{revolution}$_{\textrm{\textcolor{blue}{Fem-Sg}}}$ on himself, that he \textbf{operated}$_{\textrm{\textcolor{blue}{Fem-Sg}}}$...), the automatically parsed annotation erroneously identified the antecdent as `way' instead of `revolution'. The two missed examples reflect also the difficulty of this task for a model.

\begin{table}
  \centering
  \scalebox{.8}{
  \begin{tabular}{lllcc}
    \toprule
    \multicolumn{2}{l}{corpus} & \makecell{size \\ {\scriptsize in sentences}} & LSTMs & Transformers \\
    \midrule
    \multicolumn{4}{l}{\textit{Nonce Test Set}} \\
     & overall & 68,497 & 78.1 $_{\pm 1.2}$  & 92.6 $_{\pm 1.9}$     \\\cline{2-5}
     & 4 heuristics &32,311 &94.3 $_{\pm 1.1}$ & 98.3 $_{\pm 0.7}$\\
     & 3 heuristics & 13,222 & 80.3 $_{\pm 2.5}$& 93.5 $_{\pm 1.9}$\\
     & 2 heuristics & 8,869 & 63.2 $_{\pm 2.1}$& 89.1 $_{\pm 2.9}$\\    
     & 1 heuristic & 10,946&53.0 $_{\pm 5.1}$& 84.0  $_{\pm 3.5}$ \\
     & 0 heuristic  & 3,149&32.3 $_{\pm 11}$& 69.1 $_{\pm 4.5}$\\
    \bottomrule
  \end{tabular}}
  \caption{Accuracy achieved by LSTMs and Transformers on the \textit{nonce test set}, based on prediction difficulty \label{results_by_heuristics}}
\end{table}

\begin{figure*}
\tikzstyle{every picture}+=[remember picture,inner xsep=0,inner ysep=0.25ex]

  \begin{tabu}{cllllllllll}

  (1) &Les &  \tikz[baseline=(node1.base)]\node (node1){offres}; & \tikz[baseline=(node2.base)]\node (node2){\alert{ que} }; & Pierre &dit& que & Marie& a & \alert{acceptées} &  \\
  \rowfont{\tiny}&The & offers & that & Peter &sayss& that & Mary &has& accepted & \\
  & \multicolumn{9}{l}{The offers that Peter says Mary accepted.} \\\\\\
  
  (2) &Les &  \tikz[baseline=(node3.base)]\node (node3){disques}; &et & les &livres& \tikz[baseline=(node4.base)]\node (node4){\alert{qu'}}; & il& a & \alert{achetés} &  \\
  \rowfont{\tiny}&The & disks & and & the &books& that & he &has& bought & \\
  & \multicolumn{9}{l}{Disks and books that he has bought... } \\\\\\
  
  (3) &Les &  \tikz[baseline=(node5.base)]\node (node5){propositions}; &de & la &fédération& \tikz[baseline=(node6.base)]\node (node6){\alert{qu'}}; & il& a & \alert{faites} &  \\
  \rowfont{\tiny}&The & proposals & of & the &federation& that & he &has& made & \\
  & \multicolumn{9}{l}{The proposals of the federation that he has made... } \\\\
  \end{tabu}
  \caption{ The test set excluded the complex long distance dependencies (1) and ambiguous coordinated object noun phrase (2), but kept the prepositional phrase as antecedent cases like (3) \label{fig:ex_test_set}}
    
    \begin{tikzpicture}[overlay]
    \draw[-latex,dashed] (node2.north) to[bend right] (node1.north);
    \draw[-latex,dashed] (node4.north) to[bend right] (node3.north);
    \draw[-latex,dashed] (node6.north) to[bend right] (node5.north);
\end{tikzpicture}
\end{figure*}

\paragraph{Nonce test set} To test the extent to which the lexical or collocational information
contribute to model's performance on number agreement task, we adapted the generation procedure of
~\newcite{gulordava-etal-2018-colorless} to generate three "colorless
green idea" ~\cite{chomsky1957mouton} sentences for each original
sentence: each content word of the original sentence is replaced with
a random word from the same syntactic category (i.e.,the same POS and
morphological features). During the substitution procedure, we
excluded the word forms that appeared in the treebank with more than
one POS to make sure that the random words used are all with
unambiguous POS(e.g., {\it données} can be a plural noun(data) or
the plural past participle of verb {\it donner} (give)). To respect
the argument structure constraints, the target verb could only be
replaced by another random transitive word. So the \textit{Nonce Test Set}
retains the grammatical syntax of original sentences but are highly
semantically implausible. Table \ref{ex_control} gives an example of a nonsensical sentence converted from its original version.

\paragraph{Mirror test set} We generate a singular version of each
plural object sentence and vice versa by substituting respectively the
antecedent and target verb of each original sentence
with their opposite number form. We converted also the adjective and
pronoun modifiers of the antecedent to their opposite number form if
they are present. At the end, we got a "inverted copy" of the original
set in terms of class distribution, 35\% singular and 65\% plural compared to its original version: 65\% singular and
35\% plural. Table \ref{ex_control} gives an example of the \textit{mirror} sentence converted from its original version.

\begin{table*}
  \begin{tabular}{lll}
    \toprule
    Test sets & Examples & label \\
    \midrule
    Original& Les \textbf{offres} que le directeur a \textbf{acceptées}...  &Pl     \\
    & \textit {The offers\_\textbf{Pl} that the director has accepted\_\textbf{Pl} ...}  &   \\
    Nonce& Les \textbf{omellettes} que le professeur a \textbf{attachées}...  &Pl  \\
    & \textit {The omelettes\_\textbf{Pl} that the professor has attached\_\textbf{Pl} ...}   &  \\
    Mirror& L' \textbf{offre} que le directeur a \textbf{acceptée}  & Sg    \\
    & \textit {The offer\_\textbf{Sg} that the director has accepted\_\textbf{Sg} ...}   &  \\
    Permuted& directeur a Les que \textbf{offres} le  \textbf{acceptées} ...  &Pl   \\
    & \textit {director has The that  offers\_\textbf{Pl}  the accepted\_\textbf{Pl} ...}  &   \\
    \bottomrule
  \end{tabular}
  \caption{Examples of test sets used in original and control experiments \label{ex_control}}
\end{table*}

\section{Detailed results}
\begin{table*}
  \begin{tabular}{llllllll}
    \toprule
     &  2 tokens & 3-4 & 5-6 & 7-8 & 9-10& 11-12 & 13-14  \\
    \midrule
    LSTMs     & 73.1$_{\pm 0.9}$  & 82.9$_{\pm 1.5}$ & 78.7$_{\pm 1.2}$ & 75.9$_{\pm 0.6}$ & 74.1$_{\pm 0.3}$& 72$_{\pm 0.6}$ & 69.3$_{\pm 1.2}$ \\
    Transformers     & 88.0$_{\pm 3.0}$  & 95.1$_{\pm 1.2}$ & 92.4$_{\pm 1.6}$ & 89.7$_{\pm 1.9}$ & 87.8$_{\pm 2.2}$& 85.2$_{\pm 2.2}$ & 83.1$_{\pm 1.7}$\\
    \# examples & 1,599 & 44,012&14,945&4,799&1729&756&327\\
    \bottomrule
  \end{tabular}
  \caption{Accuracy as a function of distance (i.e. number of tokens) between the antecedent and the target verb \label{distance}}
  \centering
\end{table*}

\paragraph{Nonce Set} The detailed results on \textit{Nonce Test Set} are reported in Table \ref{results_by_heuristics}.

\paragraph{Distance} Table \ref{distance} reports the average prediction accuracy on \textit{Original Test set} as a function of distance between the antecedent and the target verb. The shortest distance (i.e. construction with only two intervening
tokens: the relative pronoun and the auxiliary verb) is more
challenging for both LSTMs and Transformers due to the attraction
effect of the auxiliary. In this non-canonical construction(1,599 examples), the embedded subject in the objective clause occurs after its predicate. Our fine-grained analysis shows that in this non-canonical case, when the number of the intervening auxiliary is different with that of the past participle verb, LSTMs' performance drops to 41.9\% and Transformers
still achieve an accuracy of 80\%, suggesting that Transformer are
more robust to resist the lure of adjacent auxiliary attractor.

\end{document}